\documentclass{article}
\usepackage{spconf,graphicx}
\usepackage{amsmath,amssymb,amsfonts}
\usepackage{algorithmic}
\usepackage{graphicx}
\usepackage{textcomp}
\usepackage{xcolor}
\def\BibTeX{{\rm B\kern-.05em{\sc i\kern-.025em b}\kern-.08em
    T\kern-.1667em\lower.7ex\hbox{E}\kern-.125emX}}

\usepackage{makecell}
\usepackage{booktabs}
\usepackage{threeparttable}

\usepackage{hyperref}

\begin{document}

\title{State-of-the-Art in Nudity Classification: \\A Comparative Analysis}

\name{Fatih Cagatay Akyon$^{1,2}$, Alptekin Temizel$^{1}$}
\address{$^{1}$Graduate School of Informatics, METU, Ankara, Turkey \\
$^{2}$OBSS AI, OBSS Technology, Ankara, Turkey}

%\author{\IEEEauthorblockN{Fatih Cagatay Akyon}
%\IEEEauthorblockA{ \\
%\textit{Graduate School of Informatics, METU}\\
%Ankara, Turkey \\
%}
%\and
%\IEEEauthorblockN{Alptekin Temizel}
%\IEEEauthorblockA{ \\
%\textit{Graduate School of Informatics, METU}\\
%Ankara, Turkey \\
%}
%}

\maketitle

\begin{abstract}
This paper presents a comparative analysis of existing nudity classification techniques for classifying images based on the presence of nudity, with a focus on their application in content moderation. The evaluation focuses on CNN-based models, vision transformer, and popular open-source safety checkers from Stable Diffusion and Large-scale Artificial Intelligence Open Network (LAION). The study identifies the limitations of current evaluation datasets and highlights the need for more diverse and challenging datasets. The paper discusses the potential implications of these findings for developing more accurate and effective image classification systems on online platforms. Overall, the study emphasizes the importance of continually improving image classification models to ensure the safety and well-being of platform users. The project page, including the demonstrations and results is publicly available at \href{https://github.com/fcakyon/content-moderation-deep-learning}{https://github.com/fcakyon/content-moderation-deep-learning}.
\end{abstract}

%\begin{IEEEkeywords}
%\end{IEEEkeywords}

\section{Introduction}
The rapid increase in user-generated content online has led to a pressing need for automated systems for filtering inappropriate and harmful content \cite{gorwa2020algorithmic} \cite{akyon2022contentmoderation}.
A recent study \cite{seyma_ben_nudity} revealed that parents have a high level of concern about the negative effects of inappropriate content in the media on their children's development and well-being, especially sexual content. In this context, the development of effective nudity classification systems has become essential for content moderation in online platforms to ensure user safety and well-being. While traditional machine learning models \cite{tofa2017inappropriate} \cite{vishwakarma2019hybrid} and convolutional neural networks (CNNs) \cite{pandey2021device} \cite{gruosso2019deep} \cite{de2019multimodal} \cite{de2019baseline}, have been widely used for nudity classification, recent transformer-based models \cite{vit} and modern CNN architectures \cite{convnext} have shown promising results in image classification tasks.

With the increasing popularity and prominence of text-to-image generation systems, artificially generated unsafe and inappropriate images have become a major concern for content moderation. This created a need for systems facilitating automated safety check procedures such as Stable Diffusion Safety checker \cite{stablediffusion} and LAION safety checker \cite{laion_safety}. Stable Diffusion safety checker is designed to prevent unsafe image generation, while LAION safety checker works by filtering out unwanted images from the training set to prevent diffusion models being trained on inappropriate images. These safety checkers demonstrate the growing importance of developing effective and accurate image classification systems for content moderation in online platforms. By improving the accuracy and effectiveness of image classification models, online communities can be better protected from harmful and inappropriate content.

However, the limitations of current evaluation benchmarks and datasets \cite{nudenet_dataset} \cite{LSPD_dataset} \cite{adult_content_dataset} have raised concerns about the effectiveness of these models in accurately detecting nudity. Therefore, in this paper, we present a comparative analysis of existing nudity classification techniques for classifying images based on the presence of nudity, with a focus on their application in content moderation. We evaluate CNN-based models, recent transformer-based models, and popular open-source safety checkers and highlight the limitations of current evaluation datasets.
The findings of this study are expected to contribute to the development of more effective and culturally-sensitive image classification systems for content moderation in online platforms.

\section{Related Work}

In this section, we provide an overview of existing research on the nudity classification from images, particularly focusing on nudity classification datasets and techniques, and image classification techniques, highlighting their key findings and contributions to the field.

\subsection{Nudity Classification Datasets}

\begin{table*}
\small
\begin{center}
\caption{List of nudity classification datasets used in this work.}
\label{table:datasets}
\vspace{0.3cm}
\begin{tabular}{ c c c c c } 
\hline
Name & Year & Task & Labels \\
\hline
LSPD \cite{LSPD_dataset} & 2022 & image classification & \makecell{porn, normal, sexy, \\ hentai, drawings} \\ 
\hline
\makecell{NudeNet \\ dataset \cite{nudenet_dataset}} & 2019 & image classification & safe, sexy, nude \\
\hline
\makecell{Adult content \\ dataset \cite{adult_content_dataset}} & 2017 & image classification & safe, adult \\
\hline
\end{tabular}

\end{center}
\end{table*}

In this work, we use the following three datasets: Adult content dataset, NudeNet, and LPSD (Table \ref{table:datasets}). Adult content dataset \cite{adult_content_dataset} is one of the earliest datasets in this field and contains two categories:  `safe' and  `adult'. NudeNet dataset \cite{nudenet_dataset} includes an intermediate label `sexy', in addition to the `safe' and `nude' labels. LSPD \cite{LSPD_dataset} consists of image and video classification branches that provide nudity and pornography detection-related annotations available to researchers upon request. LSPD includes the most detailed classification labels among these three datasets. 

\subsection{Content Moderation and Nudity Classification}

The problem of content moderation on online platforms has spurred a lot of research in recent years. One important aspect of content moderation is image classification, particularly in detecting and classifying nudity.

A system for nudity classification from images using a Mobilenetv3 image embedding model \cite{mobilenetv3} was proposed in \cite{pandey2021device}. It was designed for on-device content moderation and aimed to classify images as `containing nudity'/`not containing nudity'. In \cite{larocque2021gore}, a CNN + MLP ensemble was proposed to classify gore in images. While the study did not particularly focus on nudity detection, the approach using image embeddings from Mobilenet v2 \cite{mobilenetv2}, Densenet \cite{densenet}, and VGG16 \cite{vgg16} could be adapted for this purpose.

A CNN + SVM model for inappropriate video scene classification, including nudity was proposed in \cite{de2019multimodal}. The model is based on InceptionV3 \cite{inceptionv3} image embeddings and was trained on a private dataset. A novel method for learning scene representations from movies using a ViT-like video encoder and MLP was proposed in \cite{chen2022movies2scenes}. While the study does not focus specifically on image-based nudity classification, the ViT-like encoder used in the model can be utilized for this purpose. The proposed approach could potentially improve content moderation for online platforms by classifying video scenes based on their content rating related to nudity detection using image classification. The model was evaluated on multiple datasets, including a private dataset, and demonstrated promising results.

\section{Experimental Evaluation and Results}

\begin{table*}[!h]
    %\vspace{-0.5cm}
    \caption{Model output label mapping used in zero-shot NudeNet experiments.}
    \centering
    \small
    \begin{tabular}{c|c}
    model output labels & target NudeNet label \\
    \toprule
    normal, drawing & safe \\
    sexy & sexy \\
    hentai, porn & nude \\
    \bottomrule
    \end{tabular}
    \vspace{0.2cm}

    \label{tab:nudenet-mapping}
    %\vspace{-0.2cm}
\end{table*}

\begin{table*}[!h]
    %\vspace{-0.5cm}
    \caption{Model output label mapping used in zero-shot AdultContent experiments.}
    \centering
    \small
    \begin{tabular}{c|c}
    model output labels & target AdultContent label \\
    \toprule
    normal, drawing & safe \\
    sexy, hentai, porn & adult \\
    \bottomrule
    \end{tabular}
    \vspace{0.2cm}

    \label{tab:adultcontent-mapping}
    %\vspace{-0.2cm}
\end{table*}

\begin{table*}[!h]
    %\vspace{-0.5cm}
    \caption{Overall and label-wise test set results on the LSPD image classification dataset.}
    \centering
    \small
    \begin{threeparttable}
    \begin{tabular}{c|cccc|ccccc}
    Model & F1$_{all}$ & Acc.$_{all}$ & Prec.$_{all}$ & Rec.$_{all}$ & F1$_{normal}$ & F1$_{drawing}$ & F1$_{sexy}$ & F1$_{hentai}$ & F1$_{porn}$\\
    \toprule
    MobileNetv2 & 0.945 & 0.941 & 0.950 & 0.941 & 0.968 & 0.954 & 0.900 & 0.948 & 0.957 \\
    MobileNetv3(small) & 0.934 & 0.930 & \textbf{0.953} & 0.930 & 0.960 & 0.944 & 0.883 & 0.939 & 0.946 \\
    MobileNetv3(large) & 0.949 & \textbf{0.946} & \textbf{0.956} & \textbf{0.946} & \textbf{0.972} & \textbf{0.958} & 0.905 & \textbf{0.953} & \textbf{0.958} \\
    Inceptionv3 & \textbf{0.950} & \textbf{0.947} & \textbf{0.954} & \textbf{0.947} & \textbf{0.970} & \textbf{0.960} & \textbf{0.909} & \textbf{0.952} & \textbf{0.961} \\
    ConvNexT(tiny) & \textbf{0.952} & \textbf{0.949} & \textbf{0.954} & \textbf{0.949} & \textbf{0.971} & \textbf{0.960} & \textbf{0.913} & \textbf{0.955} & \textbf{0.962} \\
    ViT(B16) & 0.920 & 0.914 & 0.927 & 0.914 & 0.941 & 0.929 & 0.858 & 0.931 & 0.938 \\
    \midrule
    LAION Safety Checker\tnote{*} & 0.791 & 0.809 & 0.804 & 0.809 & 0.893 & 0.855 & 0.623 & 0.891 & 0.692 \\
    \bottomrule
    \end{tabular}
    \begin{tablenotes}
        \item[*]Model not trained on the given dataset.
    \end{tablenotes}
    \end{threeparttable}
    \vspace{0.2cm}

    \label{tab:lspd-result}
    %\vspace{-0.2cm}
\end{table*}

\begin{table*}[!h]
    %\vspace{-0.5cm}
    \caption{Overall and label-wise test set results on the NudeNet image classification dataset.}
    \centering
    \small
    \begin{threeparttable}
    \begin{tabular}{c|cccc|ccccc}
    Model & F1$_{all}$ & Acc.$_{all}$ & Prec.$_{all}$ & Rec.$_{all}$ & F1$_{safe}$ & F1$_{sexy}$ & F1$_{nude}$ \\
    \toprule
    MobileNetv2 & 0.882 & 0.866 & 0.911 & 0.848 & 0.938 & 0.819 & 0.882 \\
    MobileNetv3(small) & 0.857 & 0.840 & 0.902 & 0.840 & 0.938 & 0.772 & 0.863 \\
    MobileNetv3(large) & 0.882 & \textbf{0.866} & 0.918 & \textbf{0.866} & 0.946 & 0.819 & 0.882 \\
    Inceptionv3 & 0.878 & 0.862 & 0.918 & 0.862 & \textbf{0.953} & 0.801 & 0.881 \\
    ConvNexT(tiny) & \textbf{0.886} & \textbf{0.870} & \textbf{0.922} & \textbf{0.870} & \textbf{0.951} & \textbf{0.822} & \textbf{0.886} \\
    ViT(B16) & 0.839 & 0.820 & 0.892 & 0.820 & 0.920 & 0.752 & 0.844 \\
    \midrule
    LAION Safety Checker\tnote{*} & 0.882 & \textbf{0.889} & 0.881 & \textbf{0.889} & 0.928 & \textbf{0.855} & 0.863 \\
    ConvNexT(tiny)-LSPD\tnote{*} & \textbf{0.893} & 0.882 & \textbf{0.913} & 0.882 & \textbf{0.941} & 0.843 & \textbf{0.896} \\
    \bottomrule
    \end{tabular}
    \begin{tablenotes}
        \item[*]Model not trained on the given dataset.
    \end{tablenotes}
    \end{threeparttable}
    \vspace{0.2cm}

    \label{tab:nude-result}
    %\vspace{-0.2cm}
\end{table*}

\begin{table*}[!h]
    %\vspace{-0.5cm}
    \caption{Overall and label-wise test set results on the AdultContent image classification dataset.}
    \centering
    \small
    \begin{threeparttable}
    \begin{tabular}{c|cccc|ccccc}
    Model & F1$_{all}$ & Acc.$_{all}$ & Prec.$_{all}$ & Rec.$_{all}$ & F1$_{safe}$ & F1$_{adult}$ \\
    \toprule
    MobileNetv2 & \textbf{0.989} & \textbf{0.990} & \textbf{0.990} & \textbf{0.989} & \textbf{0.994} & 0.976 \\
    MobileNetv3(small) & 0.987 & 0.987 & 0.987 & 0.987 & \textbf{0.992} & 0.970 \\
    MobileNetv3(large) & \textbf{0.989} & \textbf{0.989} & \textbf{0.989} & \textbf{0.989} & \textbf{0.994} & 0.974 \\
    Inceptionv3 & \textbf{0.988} & \textbf{0.988} & \textbf{0.988} & \textbf{0.988} & \textbf{0.994} & 0.972 \\
    ConvNexT(tiny) & \textbf{0.991} & \textbf{0.991} & \textbf{0.991} & \textbf{0.991} & \textbf{0.995} & \textbf{0.980} \\
    ViT(B16) & 0.978 & 0.978 & 0.978 & 0.978 & 0.978 & \textbf{0.978} \\
    \midrule
    LAION Safety Checker\tnote{*} & 0.974 & 0.974 & 0.974 & 0.974 & \textbf{0.986} & 0.962 \\
    Stable Diffusion Safety Checker\tnote{*} & 0.939 & 0.940 & 0.944 & 0.940 & 0.927 & 0.950 \\
    ConvNexT(tiny)-LSPD\tnote{*} & \textbf{0.984} & \textbf{0.984} & \textbf{0.979} & \textbf{0.988} & 0.978 & \textbf{0.989} \\
    \bottomrule
    \end{tabular}
    \begin{tablenotes}
        \item[*]Model not trained on the given dataset.
    \end{tablenotes}
    \end{threeparttable}
    \vspace{0.2cm}

    \label{tab:ac-result}
    %\vspace{-0.2cm}
\end{table*}

The experiment setup for this study involved evaluating six different models, including MobileNetv3(small), MobileNetv3(large), Inceptionv3, ConvNexT(tiny), ViT(B16), and popular open-source safety checkers from Stable Diffusion and LAION. The models were trained and tested on three different datasets, including LSPD, NudeNet, and AdultContent, each containing unique challenges and limitations in nudity classification. The training process for all the deep learning models was performed using the Adam optimizer with a learning rate of 1e-3, a cosine scheduler with 10\% warmup, and a batch size of 256 for six epochs. The evaluation metrics used in the study included label-wise F1 score, accuracy, precision, and recall for all labels in the test sets of the datasets. Additionally, overall scores are calculated by macro averaging the label-wise scores and denoted with `all' subscript in the results.
The evaluation focuses on comparing the performance of different models in classifying images based on the presence of nudity, emphasizing their application in content moderation. 

Table \ref{tab:lspd-result} presents the overall and label-wise test set results on the LSPD dataset. Fully convolutional models, including MobileNetv3, Inceptionv3, and ConvNexT, achieved high accuracy in classifying images based on the presence of nudity, with ConvNexT(tiny) achieving the highest F1 score. ViT did not achieve good results due to slow convergence. This may be due to inductive bias present in CNN's or the limited transfer learning capability of the transformer-based models. Furthermore, the table presents the zero-shot performance of the LAION Safety Checker.

In NudeNet and AdultContent zero-shot experiments, ConvNexT(tiny)-LSPD (the ConvNexT model trained on the LSPD dataset) and LAION Safety Checker model outputs are mapped into NudeNet and AdultContent labels as given in Table \ref{tab:nudenet-mapping}, and Table \ref{tab:adultcontent-mapping}, respectively.

Table \ref{tab:nude-result} presents the overall and label-wise test set results on the NudeNet dataset. Fully convolutional models, including MobileNetv3, Inceptionv3, and ConvNexT, achieved high accuracy in classifying images based on the presence of nudity, with ConvNexT achieving the highest F1 score. The zero-shot results for the LAION Safety Checker and ConvNexT(tiny)-LSPD were also presented. ConvNexT(tiny)-LSPD outperforms LAION Safety Checker by 1.1 points in F1$_{all}$ score and by 3.3 points in F1$_{nude}$.

Table \ref{tab:ac-result} presents the overall and label-wise test set results on the AdultContent dataset. ConvNexT performs best in the supervised setting. In zero-shot setting, ConvNexT(tiny)-LSPD outperforms LAION and Stable Diffusion safety checkers by 1 and 4.5 points in F1$_{all}$, respectively.

Overall, the results on these three datasets show a marginal performance difference between different models, and performance is saturated. This observation highlights that there is a need for a new fine-grained nudity classification benchmark. In addition, the dataset labels can be combined in a multi-label setting (`safe' or `nude' per image and other sub-categories per image), which requires an additional labeling effort. Moreover, the current `sexy' category in datasets is not clearly defined and not useful in real-world settings. It is because of the fact that the `sexy' label in the LSPD dataset consists of images including `explicit nudity', `bikini', `lingerie', and `cleavage' concepts, while the `sexy' label in NudeNet does not contain images including `explicit nudity'. Furthermore, in real-world applications, it is very important to know the level of nudity and sexiness. For instance, `nudity' without explicit exposure of sexual body parts can be suitable in some cases, or sexiness with `cleavage' can be safe, while `lingerie' would be inappropriate. To overcome these issues of the current datasets, more detailed multiple hierarchical labels are required per image.

\section{Conclusion}
Evaluation of six different models on three different datasets shows that fully convolutional models, such as MobileNetv3, Inceptionv3, and ConvNexT, perform better than transformer-based models like ViT in nudity classification. The limited transfer learning capability of transformer-based models may be a contributing factor to their lower performance. Furthermore, the study highlights inferior zero-shot performance of popular safety checkers from Stable Diffusion and LAION and presents better alternatives. Overall, the study emphasizes the need for continual improvement of image classification models to ensure the safety and well-being of platform users. Additionally, there is a need for a new fine-grained nudity classification benchmark that can better represent the real-world challenges of nudity detection in online platforms. The project page is available at \href{https://github.com/fcakyon/content-moderation-deep-learning}{https://github.com/fcakyon/content-moderation-deep-learning} and will be updated with further demonstrations and results.

\bibliographystyle{IEEEbib}
\bibliography{refs}

\end{document}